\documentclass[letterpaper, 10 pt, conference]{ieeeconf}  % Comment this line out if you need a4paper

\IEEEoverridecommandlockouts                              % This command is only needed if 
\usepackage[compact]{titlesec}
\titlespacing{\section}{0pt}{0.3ex}{1ex}
\titlespacing{\subsection}{0pt}{0.1ex}{0ex}
\titlespacing{\subsubsection}{0pt}{0.1ex}{0ex}

\usepackage{graphics} % for pdf, bitmapped graphics files
\usepackage{amssymb}  % assumes amsmath package installed
\usepackage{comment} % this is added by me
\usepackage[demo]{graphicx} % added by mehttps://www.overleaf.com/project/61352e7e15a01d576d34069b
\usepackage{subcaption} % this is added by me, allowed to use?

\title{\LARGE \bf
PixSelect: Less but Reliable Pixels for Accurate and Efficient Localization
}

\author{Mohammad Altillawi$^{1}$% <-this % stops a space
\thanks{$^{1}$PhD student at the Computer Vision Center/Universitat Aut\`onoma de Barcelona. He conducts his research at Munich Research Center/Huawei.
        {\tt\small mohammad.altillawi@Huawei.com}}%
        }

\begin{document}

\maketitle
\thispagestyle{empty}
\pagestyle{empty}
    
%%%%%%%%%%%%%%%%%%%%%%%%%%%%%%%%%%%%%%%%%%%%%%%%%%%%%%%%%%%%%%%%%%%%%%%%%%%%%%%%

\begin{abstract}

Accurate camera pose estimation is a fundamental requirement for numerous applications, such as autonomous driving, mobile robotics, and augmented reality. In this work, we address the problem of estimating the global 6 DoF camera pose from a single RGB image in a given environment. Previous works consider every part of the image valuable for localization. However, many image regions such as the sky, occlusions, and repetitive non-distinguishable patterns cannot be utilized for localization. In addition to adding unnecessary computation efforts, extracting and matching features from such regions produce many wrong matches which in turn degrades the localization accuracy and efficiency. Our work addresses this particular issue and shows by exploiting an interesting concept of sparse 3D models that we can exploit discriminatory environment parts and avoid useless image regions for the sake of a single image localization. Interestingly, through avoiding selecting keypoints from non-reliable image regions such as trees, bushes, cars, pedestrians, and occlusions, our work acts naturally as an outlier filter. This makes our system highly efficient in that minimal set of correspondences is needed and highly accurate as the number of outliers is low. Our work exceeds state-of-the-art methods on outdoor Cambridge Landmarks dataset. With only relying on single image at inference, it outweighs in terms of accuracy methods that exploit pose priors and/or reference 3D models while being much faster. By choosing as little as 100 correspondences, it surpasses similar methods that localize from thousands of correspondences, while being more efficient. In particular, it achieves, compared to these methods, an improvement of localization by 33\% on OldHospital scene. Furthermore, It outstands direct pose regressors even those that learn from sequence of images.

\end{abstract}

\begin{figure}[t]
\centering

    \begin{subfigure}{\linewidth}
  \centering
    \includegraphics[scale=0.195]{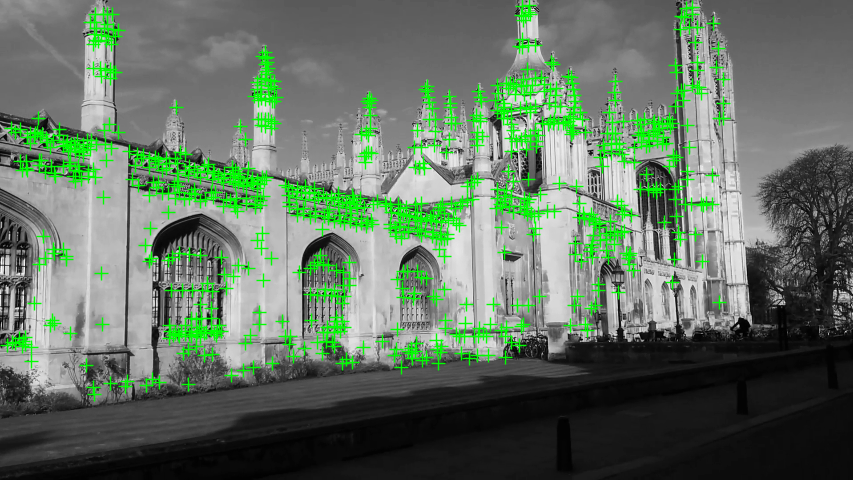}
    \includegraphics[scale=0.195]{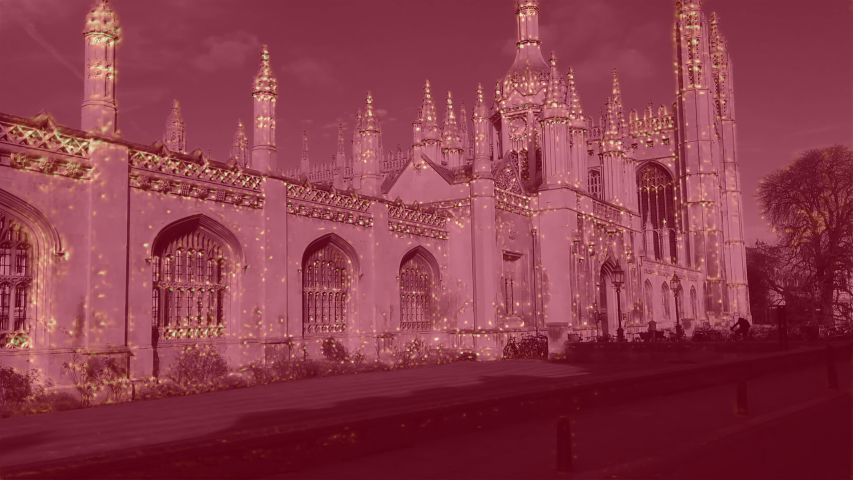}
  %\caption{}
  \end{subfigure}\smallskip %\par\medskip
  
    \begin{subfigure}{\linewidth}
  \centering
    \includegraphics[scale=0.195]{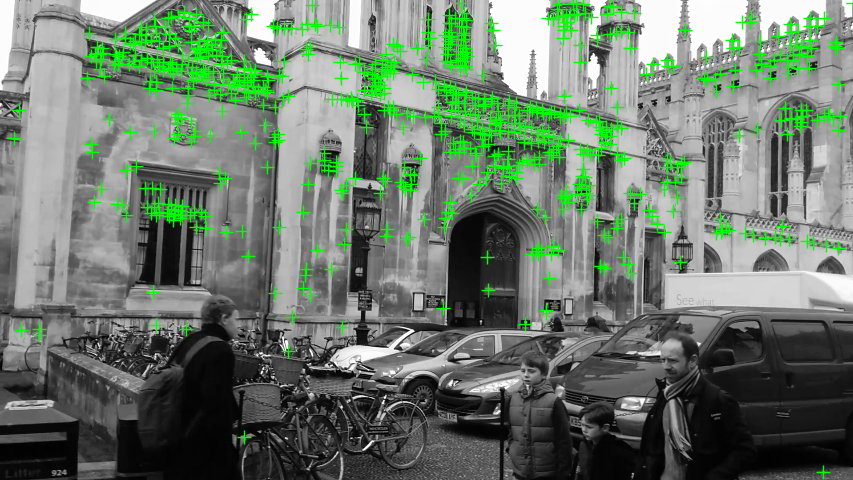}
    \includegraphics[scale=0.195]{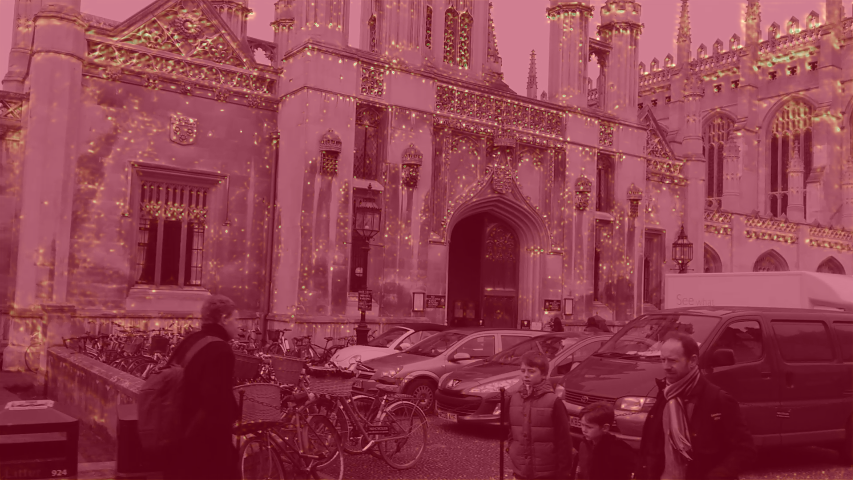}
  %\caption{}
  \end{subfigure}\smallskip %\par\medskip
  
     \begin{subfigure}{\linewidth}
  \centering
    \includegraphics[scale=0.195]{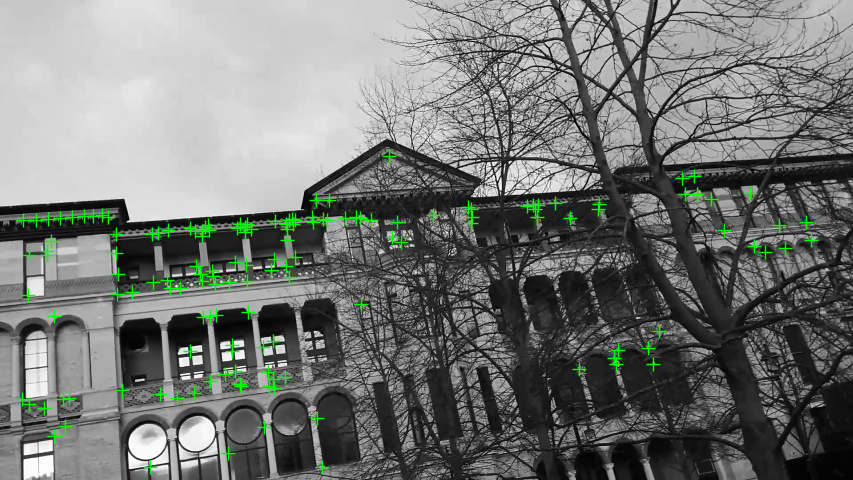}
    \includegraphics[scale=0.195]{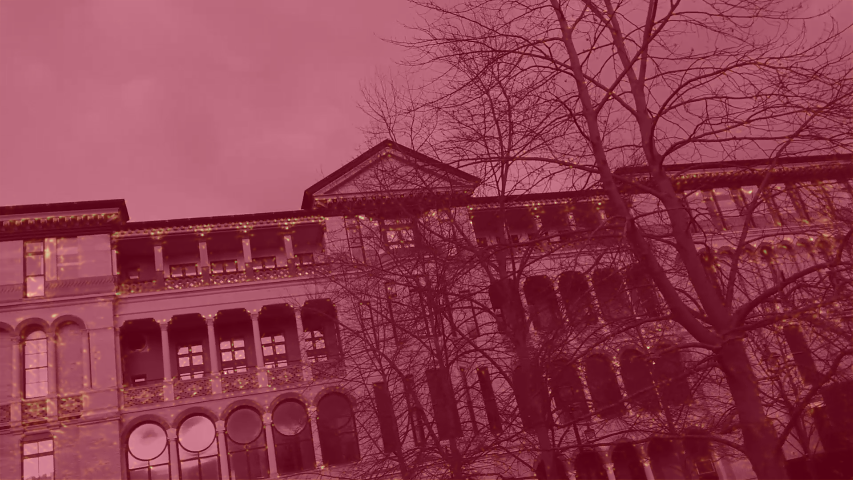}
  %\caption{}
  \end{subfigure} %\par\medskip

  \caption{For visual localization, not all image regions are of equal importance. Our method selects 2D keypoints from image regions that are discriminatory and very useful for localization. The left side images show the valid keypoints obtained by our method for the sake of localization. The right side images are the corresponding predicted heatmaps where good image regions are highlighted in yellowish-green (best seen in colors). Interestingly, our method was able to emphasize reliable regions for localization and avoid other non relevant regions such as the trees, the cloudy sky, repetitive plane structures like walls, in addition to pedestrians, and cars. We draw your attention in particular to the bottom image and the corresponding heatmap, where it can be observed that our method was able to distinguish the trees from the structure, avoid selection from the tree branches while emphasizing the background structure. It also ignored the reflective rounded windows.}
  \label{teaser}
\end{figure}

%%%%%%%%%%%%%%%%%%%%%%%%%%%%%%%%%%%%%%%%%%%%%%%%%%%%%%%%%%%%%%%%%%%%%%%%%%%%%%%%
\vspace{0.2cm}
\section{INTRODUCTION}

Camera localization has enabled robots and cars to navigate in areas where other localization sensors such as GPS may fail. Due to delays or blockage of satellite signals, GPS becomes less reliable affecting the localization accuracy.     
Camera based global localization has been addressed by classical structure based methods \cite{structr_1, structr_2, activesearch} as well as by deep learning based approaches \cite{posenet, posenet+, poselstm, mapnet, attloc, vipr, vidloc, vlocnet, dsac, dsac++, dsacstar}.

Classic structure based methods \cite{structr_1, structr_2, activesearch} estimate a pose by first extracting 2D features from query images then matching these features to the 3D points in a 3D scene model to form a set of 2D-3D correspondences which are in turn used by a PnP \cite{pnp, epnp} algorithm in a RANSAC \cite{ransac} framework to estimate a 6 DoF pose. 
Structure based methods obtain a fine pose localization. However, in online deployment, they rely on a 3D model which has a memory requirement that increases with the increase in localization scope. The descriptor matching step to obtain correspondences is expensive and time consuming procedure. In addition, the obtained correspondences are noisy and the number of outlier grows high with the growth of the model demanding growth in the runtime of RANSAC thus the localization runtime.

The recent advances in deep learning have inspired many works to utilize these advances for the benefit of localization. One direction of these works is to deploy a deep network to map input image or sequence of images to a pose \cite{posenet, posenet+, poselstm, mapnet, attloc, vipr, vidloc, vlocnet}. The common practice of these direct pose regressors is to encode the image in a latent feature vector out of which pose is regressed as two separate units which are translation and rotation. Usually, learning this direct mapping starts by transferring knowledge learned from a large classification dataset then finetuning it on the task of pose regression. The proposals of this art aimed to constraint the latent feature vector to encode useful information from the image or stream of images for the benefit of localization. Compared to structure based methods, they are faster with lower memory print. However, their localization accuracy is much lower.

Following the advances of deep learning, some works \cite{Shotton2013SceneCR, dsac, dsac++, dsacstar} took an approach different from direct pose regressors. Instead of regressing the pose, the 3D scene is regressed to form set of 2D-3D correspondences out of which a pose is calculated. In these works, every part of the image is considered valuable for localization and correspondences are obtained from every grid of the image. These works achieve the state of art localization results.

Our approach proposes a hybrid method that deploys deep learning as well as the classic geometric algorithms for pose estimation. While we utilize the learning methods to learn the 3D scene, specifically the discriminatory parts of the scene, we keep the geometric pose estimation to the classic well established algorithms mainly the PnP algorithm \cite{pnp, epnp} and RANSAC \cite{ransac} scheme.

The well established algorithms work off the shelf and independent of the scene. Though the pros that it brings such as the improved sensitivity to outliers or the enabling of end-to-end training, learning these algorithms may overfit them, in a form or another, to the specific learned scene and may hinder their generalizablility. Rather than utilizing deep learning to learn PnP or RANSAC, we keep the geometric pose estimation to the classic algorithms. Instead, we exploit deep learning to learn suitable features for localization. Following along, this work involves deep learning to simultaneously learn two tasks. The first is the 3D scene, where the camera is to be localized. The second is where to select features that are beneficial for localization. As thus, we do not consider every image region valuable for localization and avoid regions like the sky, clouds, repetitive patterns such as walls, ground, pavements, grass spaces, trees, and occluded areas. Refer to Fig. \ref{teaser} for some examples of our learned selection of reliable image regions.
 Our main contributions can be summarized as follows:
 \begin{itemize}
     \item We present a new hybrid camera re-localization pipeline that combines deep learning for learning the 3D scene and good localization features and the classic geometric algorithms for estimating the global pose. The proposed approach acts by nature as an outlier filter. It is efficient, accurate, and surpasses state of art methods on Cambridge Landmarks dataset \cite{posenet}. 
     \item We present the first approach that learns 2D keypoints and 3D scene coordinates in one framework.
     \item We present a keypoints detector that can avoid selection from non discriminatory and occluded areas.
 \end{itemize}

\begin{figure*}[h]
\centering
\includegraphics[scale=0.28]{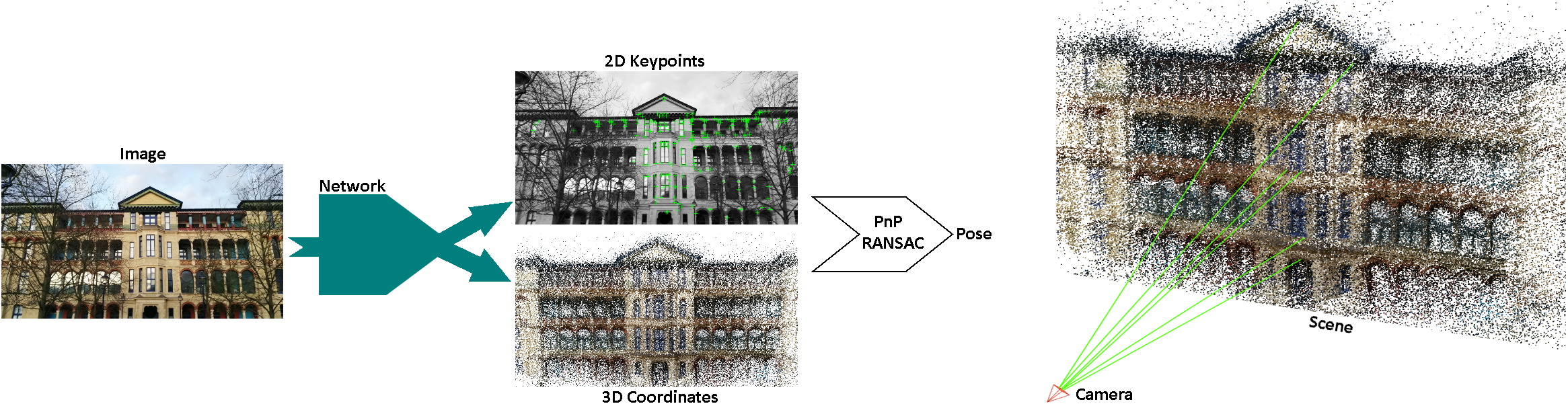}
  \caption{A sketched representation of our pipeline: The processing flow is left to right. For a given input image, a deep network produces two outputs. The first is a set of reliable 2D keypoints. Here we show already the valid 2D keypoints overlayed on the image. The second is a set of 3D coordinates relative to a global coordinate system. Here we show an illustrative example. Correspondences are then obtained directly by selecting the 3D coordinates that correspond to the valid 2D keypoints. The set of selected 2D-3D correspondences are then used to obtain a pose using Perspective-n-Point \cite{pnp} in a RANSAC scheme \cite{ransac}. The most right sketch shows the pose computation step of the camera relative to a scene where the green lines describe the 2D-3D correspondences. For a description of the network architecture, refer to section \ref{arch}.}
  \label{sketch}
\end{figure*}

\section{Related Work}
In the following, we discuss the related work for solving camera global re-localization.

%\subsection{Pose Regression}
\textbf{Pose Regression:}
Pose regressors learn the mapping function from either a single input image \cite{posenet, posenet+, poselstm} or sequence of images \cite{vidloc, mapnet, vlocnet, attloc, vipr} into 6 Dof Pose. A network encodes the input into a latent feature out of which the pose is regressed. Different learning strategies are followed to constraint the latent vector either by LSTMs \cite{poselstm}, relative pose information \cite{vidloc, mapnet, vlocnet, vipr}, 3D model \cite{posenet+} or attention \cite{attloc}. As the mapping is direct, the execution time is fast. However, the common practice is that these methods model the geometric pose estimation problem as a regression task. In contrast, we keep the pose estimation task to the classic algorithms \cite{pnp, epnp} and argue that through pose regression, 3D geometric information cannot be utilized in proper way for pose estimation resulting in inaccurate poses.

%\subsection{Sparse feature matching/alignment}
\textbf{Sparse feature matching/alignment:}
Active Search \cite{activesearch} provides a prioritization scheme to establish 2D-3D matches and terminates correspondence search once an enough number of matches is found. Opposite to that, our work obtains matches directly by regressing for the pixels of interest, their 3D coordinates in a global reference frame bypassing the computationally expensive step of descriptors matching. Furthermore, Active Search requires the reference 3D model of the target scene at inference. By learning the scene, our proposal predicts the scene at inference avoiding the necessity of a model that demands more memory with the increase of localization target area. Saving only the weights of the deep network keeps a constant memory footprint. PixLoc \cite{pixloc} utilizes a query image, reference images, pose priors, and a reference 3D SFM model to learn good features for localization. PixLoc \cite{pixloc} can improve sparse feature matching by refining keypoints and poses and localizes in large environment given a pose prior. From a simpler training method, our proposal learns better features for localization. Their work recognizes the sky and clouds in addition to repetitive non discriminatory patterns as useful for localization. In contrast, these regions are down-weighted by our work. Without relying on reference 3D model or pose priors, our work obtains lower localization median errors on Cambridge landmarks dataset \cite{posenet} while being more efficient. PixLoc \cite{pixloc} extracts features from an image in 100 ms and may require up to one second to localize an image \cite{pixloc}. 

While our keypoints detector could be paired with off-the-shelf descriptors (hand-crafted or learned) and utilized for reconstructing the environment and for sparse feature matching based localization, we explore this direction in future works. We focus in this work on global camera re-localization from a single image.

\textbf{Scene Regression:}
Rather than utilizing deep learning to learn pose regression from image encoded features, other works, similar to ours, learn the 3D scene to form a set of 2D-3D correspondences for localization. DSAC \cite{dsac} proposed a method to make RANSAC differntiable and applied the work on camera localization. The differentaible RANSAC proposed by DSAC is specific to the dataset and do not generalize to other scenes. In contrast to learned RANSAC, off-the shelf RANSAC \cite{ransac} frameworks, which we put to use, are not tight to a specific scene. Two followup works, DSAC++ \cite{dsac++} and DSAC* \cite{dsacstar} improved, among other things, the localization accuracy of DSAC \cite{dsac}. They consider every pixel of the image valuable for localization. Thus, they select thousands of correspondences to estimate a reliable pose hypothesis. In contrast to DSAC and followers, our work avoids selecting pixels from non-discriminatory image regions and occlusions. Our work estimates the reliability of an image pixel for localization. This allows our work to avoid regions which are source of outliers, prioritize correspondences, and select lower number of keypoints.
%  Our work can localize from as low as 100 correspondences.

\begin{table*}[t]
\caption{Comparison against State-of-the-art visual localization methods on the Cambridge Landmarks dataset\cite{posenet}. We report the median translation and rotation errors.}
\label{outdoor}
\begin{center}
\begin{tabular}{c c c c c c c c c}
\hline
Methods & King’s College & Old Hospital & Shop Facade & St Mary’s Church\\
\hline
%PoseLSTM \cite{poselstm}& 0.99 m, 3.65°&	1.51 m, 4.29°& 1.18 m, 7.44°& 1.52 m, 6.68°\\

PoseNet ++ \cite{posenet+}&	0.88m, 1.04°&	3.20m, 3.29°& 0.88m, 3.78°& 1.57m, 3.32°\\

VLocNet \cite{vlocnet}&	0.836m, 1.419°&	1.075m, 2.411°& 0.593m, 3.529°& 0.631m, 3.906°\\
\hline

DSAC ++ \cite{dsac++}& 0.18m, 0.3° & 0.20m, \textbf{0.3°} & 0.06m, 0.30° & 0.13m, 0.4°\\

DSAC* \cite{dsacstar}&	0.15m, 0.3°& 0.21m, 0.4°& \textbf{0.05m}, 0.3°& 0.13m, 0.4°\\
\hline
ActiveSearch \cite{activesearch}&	0.42m, 0.55°&	0.44m, 1.01°& 0.12m, 0.40°&	0.19m, 0.54°\\

PixLoc \cite{pixloc}& \textbf{0.14m}, \textbf{0.24°}& 0.16m, 0.32°& \textbf{0.05m}, \textbf{0.23°}& 0.10m, \textbf{0.34°}\\
\hline
\textbf{Ours} & \textbf{0.14m}, 0.34°&	\textbf{0.14m}, 0.5° &	0.06m, 0.5°&	\textbf{0.09m}, 0.46°\\
\hline
\end{tabular}
\end{center}
\end{table*}

\section{Method}
\textbf{Overview}: The proposed work exploits deep learning to simultaneously select the relevant 2D keypoints from the input image and estimate the corresponding 3D global coordinates. The selected set of 2D-3D correspondences are passed to a PnP solver \cite{pnp} in a RANSAC \cite{ransac} framework to obtain a pose hypothesis. The proposed work is sketched in Fig. \ref{sketch}. On one side, mapping the pixels of the image to 3D coordinates enables our method to form direct correspondences and to avoid the expensive step of matching features from image to a 3D model. On the other side, not all pixels of a given image are reliable and suitable for localization. Thus, the proposed method selects the 2D-3D correspondences that are estimated to be distinguishable keeping in mind to choose a minimalistic set of keypoints aiming for an efficient and accurate localization.

\textbf{Motivation}: In fact, many regions of a query image such as the very far sky and clouds, repetitive and non-discriminatory regions like similarly looking walls, pavements, streets, trees, and bushes, in addition to occlusions are irrelevant for localization. In the context of 3D scene regression, these repetitive patterns look the same but they belong to different locations of the environment with different target 3D coordinates. Learning the 3D coordinates of these regions confuses the network and may result in improper predictions. In the context of features matching, either from image to image or image to 3D model, these regions possess big sources of outliers. As the image patches of these regions looks similar, so are the descriptors of these patches. Matching keypoints from the mentioned regions is usually an added effort which results in few matches and a lot of outliers. Similarly are the occluding areas. Avoiding these regions not only reduces the outliers, but also makes localization efficient and more accurate as we show in the sections \ref{ours_as_outlier} and \ref{ours_as_reliable}.

% \subsection{Architecture}
\subsection{Tasks Description}
\textbf{The 2D keypoints:} The network estimates a heatmap to infer about discriminative regions of the image (refer to Fig. \ref{teaser} for visualizations of samples of predicted heatmaps). With different confidence values given for each pixels, those with the highest activations are selected together with their corresponding 3D coordinates for pose estimation. 
For selecting reliable keypoints from a single image, we exploit an effective concept from sparse 3D models. A sparse 3D model can be created by a structure from motion method \cite{sfm} by triangulating the matched keypoints from different views. The idea is that, since these keypoints are already matched from different views and passed the verification steps of SfM, they most likely belong to discriminatory regions in the image and are reliable. To the best of our knowledge, our work is the first to exploit the concept of triangulated features from structure from motion to directly learn reliable and discriminatory regions from a single image (Fig. \ref{teaser}).

In this context, the network learns to predict, from a single image, a heatmap with the most activated pixels are the ones corresponding to projections from a sparse 3D model. Given a 3D model, for every training image $I$, a reference heatmap $Z$ is created by projecting the visible 3D points (that are observed from the camera with a pose corresponding to image $I$) into the image. The projected pixels are assigned probabilities of one in the reference heatmap. The 2D keypoints branch is trained by maximizing the cosine similarity between the predicted heatmap $\hat{Z}$ and the target heatmap $Z$. For an input image $I$ of size $H \times W$, the loss can be formulated as

\begin{equation}
L_{sim}(I) = 1 - \frac{1}{|P|} \sum_{p\in P}^{}cosim(\hat{Z}[p], Z[p])
\end{equation}

where $P = { \{p\}} $ is the set of the $N \times N$ corresponding patches between $Z$ and $\hat{Z}$, $Z[p] \in \mathbb{R}^{N^2}$ is the flattened $N \times N$ patch $p$ extracted from $Z$, similarly for $\hat{Z}[p]$.

\textbf{The 3D coordinates:} The network estimates the 3D coordinates of the pixels relative to a global coordinate system. However, only the pixels chosen from the 2D keypoints heatmap are used for learning the 3D scene. That is, the loss is applied only to the selected set of 2D Keypoints and corresponding 3D points. For learning the 3D global coordinates, two losses are applied. The first minimizes the difference between the selected 2D keypoints and the corresponding projected 3D points and defined as:

\begin{equation}
L_{rep}(I) = \frac{1}{|M|} \sum_{i}^{M} \parallel  x_i - \pi(k, t, R, \hat{X}_i)\parallel_2,
\end{equation}

where $\hat{X} = { \{\hat{X}_i, ..., \hat{X}_M\}}$ is the set of the predicted 3D coordinates that directly corresponds to the set of selected 2D keypoints $x = { \{x_i, ..., x_M\}}$, $M$ is the number of selected correspondences, $\pi$ is the projection function that projects the estimated 3D point $\hat{X}_i$ into image $I$ using the intrinsic parameters matrix $k$, the groundtruth extrinsic: translation $t$, rotation $R$ of camera $I$.
The second loss minimizes the difference between the predicted and reference 3D scene coordinates. This is formulated as:

\begin{equation}
L_{3D}(I) = \frac{1}{|M|} \sum_{i}^{M} \parallel  \hat{X}_i - X_i \parallel_2,
\end{equation}

where $X = { \{X_i, ..., X_M\}}$ is the set of the groundtruth 3D coordinates that corresponds to the set of predicted 3D coordinates $\hat{X} = { \{\hat{X}_i, ..., \hat{X}_M\}}$. The total loss is then the weighted sum of the three mentioned losses:
\begin{equation}
L_{all}(I) = \lambda_{sim}L_{sim}(I) + \lambda_{rep}L_{rep}(I) + \lambda_{3D}L_{3D}(I),
\end{equation}

where $\lambda_{sim}$, $\lambda_{rep}$, and $\lambda_{3D}$ are the weighting factors for the loss terms.
\begin{table*}[t]
\caption{Median position errors on Cambridge Landmarks of our work against Neural guided RANSAC \cite{ngransac}. The results show that the correspondences obtained by our work incur low number of outliers. It acts naturally as an outlier filter.}
\label{outliers_vs_ngransac}
\begin{center}
\begin{tabular}{c c c c c}
\hline
Method & King’s College & Old Hospital & Shop Facade & St Mary’s Church\\
\hline

Neural guided RANSAC \cite{ngransac}&	\textbf{0.126m} & 0.219m & \textbf{0.056m} & 0.098m\\

Ours &	0.137m & \textbf{0.14m} & 0.06m & \textbf{0.093m}\\
\hline
\end{tabular}
\end{center}
\end{table*}

\section{Experiments and Evaluation}

\subsection{Datasets}
Following previous works, we conduct our experiments on Cambridge Landmarks dataset \cite{posenet}. Cambridge landmarks is an outdoor relocalization dataset that contains RGB images of six large scenes, each covers a landmark of several hundred or thousand square meters in Cambridge, Uk. The provided reference poses are reconstructed from structure from motion. The authors provide the train and test splits. We use the provided SFM models to obtain the groundtruth 3D points for each image. Following previous works \cite{dsac++, dsacstar, pixloc}, we do not conduct experiments on the sixth scene, the street landmark as the provided reconstruction is of poor quality. For the StMarysChurch scene, we removed some images from the training set which include solely trees or bushes with low number of reference 3D points.

\subsection{Architecture and Setup} \label{arch}
The architecture is compromised of a down-sampling encoder out of which two up-sampling branches stem out. To illustrate, we adapt the network architecture of DispNet \cite{dispnet}. We keep the contractive part of DispNet and replicate the expanding part to have two branches. In addition, we remove the side predictions and keep a final layer for each expanding branch. The 3D coordinates branch output 3 channels, each for one of the X, Y, and Z coordinates. The 2D keypoints branch output a single channel heatmap. The output is normalized to be in range [0, 1]. We keep the ReLU \cite{relu} as non linearity activation for the 2D keypoints branch but we change it to ELU \cite{elu} for the nonlinearity between the 3D coordinates branch layers.

The proposed architecture is modular in that each branch can act as a standalone work. Each branch can be trained on its own then they can be combined in one training setup or the whole architecture can be trained together from scratch. Since both branches are trained from the same signal (set of 3D points) and to facilitate the training and avoid hard fine-tuning, we train first the 3D branch with the $L_{3D}$ loss for 400k iterations. After that the additional losses can be introduced for roughly 200k iterations. For this setting, the weighting factors are 1, 0.1, 1 for $\lambda_{sim}$, $\lambda_{rep}$, and $\lambda_{3D}$ respectively. We use the Adam optimizer with $\beta_1 = 0.9$, $\beta_2 = 0.999$, and $\epsilon = 10^{-8}$, weight decay of $5\times10^{-4}$, and a learning rate of $10^{-4}$.

Input images are rescaled to 480 px height and normalized by mean and standard deviation. During training, data augmentations are applied on the fly. This includes color jittering, random scaling in the range [2/3,3/2], and random in-plane rotation in the range [-30°, 30°]. The groundtruth poses are adjusted accordingly.
For keypoints selection, we apply non maximum suppression and choose keypoints with confidence scores higher than a threshold and respectively their corresponding 3D coordinates. We use the PnP-RANSAC implementation of OpenCV \cite{opencv_library} for pose estimation.

\subsection{Baselines}
We compare our work against the deep learning pose regressors \cite{posenet+, vlocnet} approaches that exploit single as well as sequence of images and show that methods that compute the pose based on classic algorithms such as PnP and RANSAC exploit the scenery and its geometry better than the pose regressors. We compare our work against the works that exploited the 3D scene geometry. Specifically, we compare against the 3D scene learning methods of DSAC++ \cite{dsac++} and DSAC* \cite{dsacstar} and show that we obtain lower localization error from much less number of correspondences. Furthermore, we put our work in comparison against works that exploited the reference 3D SFM model including the notable classic structure based method, the Active Search \cite{activesearch} and the recent work of PixLoc \cite{pixloc} that also needs pose priors and show that, without exploiting the groundturth 3D point cloud of the scene or pose priors, that our method obtains lower localization errors and runs faster. We further compare the proposed method against Neural guided RANSAC \cite{ngransac} and show that we obtain comparable or lower localization errors indicating that our obtained correspondences include low number of outliers. We further prove that our proposal acts naturally as an outlier filter with an ablation study. For listing the results of the baselines, we report the results mentioned by their publications. For ActiveSearch \cite{activesearch}, we list the results reported by \cite{posenet+, dsac++, dsacstar}.

%\section{Results and Analysis} \label{results}

\subsection{Comparison of Localization Errors Against State of Art Methods} \label{errors}

\begin{figure}[h]
\centering
  \begin{subfigure}{\linewidth}
  \centering
    \includegraphics[scale=0.195]{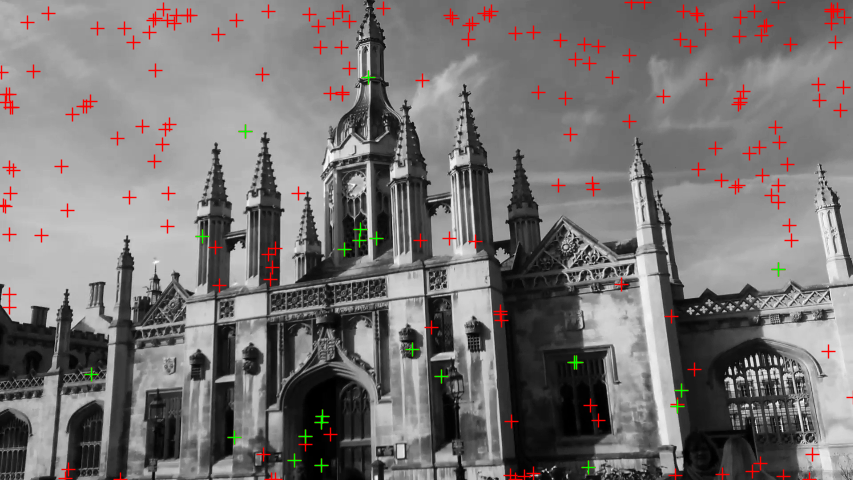}
    \includegraphics[scale=0.195]{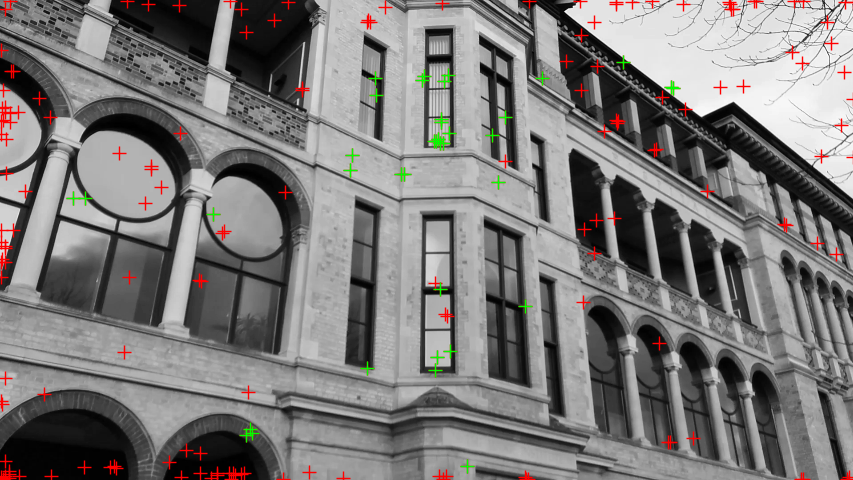}
  %\caption{}
  \end{subfigure}\smallskip %\par\medskip
  
  \begin{subfigure}{\linewidth}
  \centering
    \includegraphics[scale=0.195]{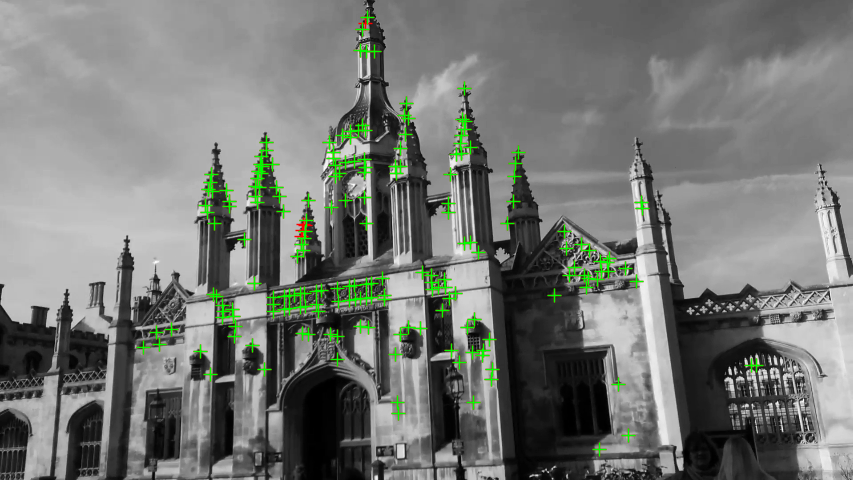}
    \includegraphics[scale=0.195]{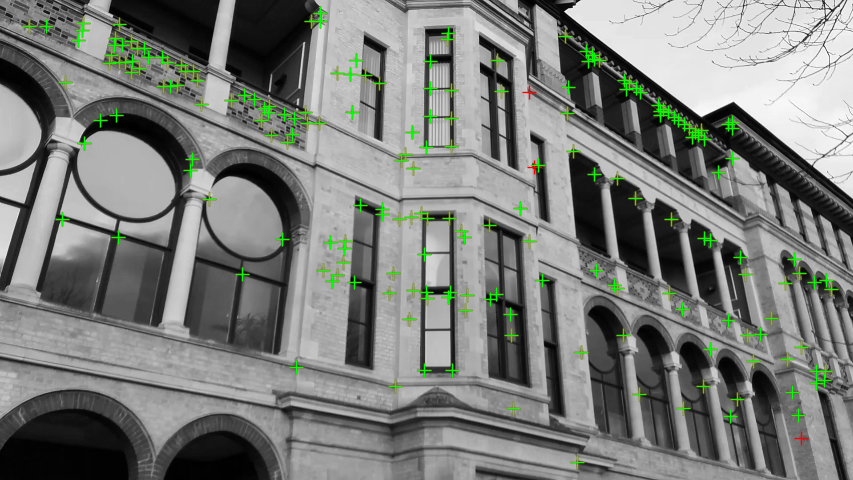}
  %\caption{}
  \end{subfigure} %\par\medskip

  \caption{Importance of selected Keypoints: The upper images show keypoints that are predicted with lower scores (set 2 of Tab. \ref{ablation_outliers}). The lower images shows keypoints that are predicted with higher scores (set 1 of Tab. \ref{ablation_outliers}). Green color denotes inliers, that is the set of keypoints that resulted in the pose hypothesis from RANSAC. Outliers are denoted by red. Left: King’s College, right: Old Hospital.}
  \label{important_pts}
\end{figure}

Tab. \ref{outdoor} reports median localization errors, as translation and rotation errors, on Cambridge landmarks dataset. Compared to DSAC++ \cite{dsac++} and DSAC* \cite{dsacstar}, the proposed work obtains lower median translation error on King’s College, Old Hospital, and St Mary’s Church and almost similar results on Shop Facade. DSAC++ \cite{dsac++}  and DSAC* \cite{dsacstar} learn the 3D scene and pass a few thousands of 2D-3D correspondences to a differentiable RANSAC to estimate the pose. In our work, we obtain much less number of correspondences. While the mentioned works select keypoints that cover the whole image, our selection is more cautious and avoids useless areas and occlusions. Figures \ref{teaser} and  \ref{important_pts} show samples of the keypoints selections of our work. The number of selected keypoints relies on how discriminatory the image regions are.
The more non-discriminatory areas (sky, similar repetitive patterns such as road, green spaces, ...) exist in the image, the lower the number of selected correspondences.
These are in average 600, 250, 130, 500 correspondences obtained from King’s College, Old Hospital, Shop Facade, and St Mary’s Church respectively.
Similarly, our work obtains lower translation error than ActiveSearch \cite{activesearch} and lower or comparable to PixLoc \cite{pixloc} results. Both works localize against a groundtruth 3D SFM model which we don't utilize at inference. PixLoc also requires a pose prior.
%Furthermore, our work obtains correspondences directly saving time by avoiding matching the keypoints against the 3D point cloud.

We further include results of pose regressors that learned from single image (PoseNet++ \cite{posenet+}) or sequence of images (VlocNet \cite{vlocnet}). Compared to Pose regressors, our method obtains much lower localization errors. PoseNet++ incorporates 3D geometric information of the scene in the loss as additional constraint for learning features suitable for localization. The work; however, treats the geometric pose estimation as a regression problem. VlocNet \cite{vlocnet}, benefits from learning from sequence of images to constraint the pose regression and obtains lower localization errors than PoseNet++ \cite{posenet+}. However, the localization errors are still higher than other works which exploited, in some form or another, 3D information for pose estimation at inference like DSAC++, DSAC*, ActiveSearch, PixLoc, and ours. Specifically, works that treated the pose geometric estimation problem using the classic computer vision methods by relying on 2D-3D correspondences obtain more accurate pose estimation than obtaining a pose by regression. This comparison was in part a motivation for us to obtain the pose using classic algorithms and confirms the observations of \cite{limitations} by further comparing against new works.

\subsection{Our Work as Outlier Filter} \label{ours_as_outlier}
We compare our work against Neural guided RANSAC, NG-RANSAC \cite{ngransac}. NG-RANSAC presents an extension to the classic RANSAC algorithm by training a network to guide RANSAC hypothesis sampling through which outliers are low weighted while inliers are predicted with higher weights. They evaluate the work on Cambridge scenes for camera localization by combining the architecture of DSAC* \cite{dsacstar} for scene regression with NG-RANSAC \cite{ngransac} for guided hypothesis sampling. Tab. \ref{outliers_vs_ngransac} lists the localization errors of NG-RANSAC and our work. By utilizing reliable keypoints, we obtain comparable results. Specifically, we achieve the best localization on the Hospital scene. In this scene, our work succeeds to avoid selecting correspondences that fall in occluded areas, cars, trees, bushes, repetitive plane areas, and the reflective windows. This emphasizes that by learning to avoid these areas, large sources of outliers are avoided thus the localization accuracy is improved.

\subsection{Importance of Selecting Reliable Keypoints} \label{ours_as_reliable}
The proposed method localizes by selecting a set of distinguishable correspondences. The selected keypoints lie in discriminatory parts of the image. This step helps selecting a minimalistic set of correspondences while incuring low level of outliers. To show the importance of the selected keypoints, we compare the pose localization errors by selecting two sets of correspondences, each has 200 correspondences. The first set is selected from the points that are predicted with high scores (above 0.7), while the other set is compromised of points that were suppressed by the non maximum suppression (predicted with score of 0.4). We run OpenCV PnP-RANSAC \cite{opencv_library} implementation with 100 iterations and a re-projection error of 3 px. Results are listed in Tab. \ref{ablation_outliers}. Furthermore, Fig. \ref{important_pts} provides visualization for the selected keypoints of the two sets.

\begin{table}[h]
\caption{Localization median errors for two sets of correspondences. Set 1: The 200 correspondences of high confidence. Set 2: 200 correspondences with lower confidence. Refer to Fig. \ref{important_pts} for visual feedback.}
\label{ablation_outliers}
\begin{center}
\begin{tabular}{ c c c}
\hline
Scene &	Set 1 & Set 2\\
\hline
King’s College &	0.15m, 0.44° &	0.43m, 1.5° \\

Old Hospital &	0.15m, 0.5° &	0.45m, 1.6°  \\
\hline
\end{tabular}
\end{center}
\end{table}

As can be observed, the set of keypoints that our method selects, belong to reliable areas. In most cases, trees, similarly looking pavements, streets, refelective windows, and sky are avoided. This has helped the network to learn the scene that corresponds to the reliable regions and predict reliably the corresponding 3D coordinates. In contrast, the already mentioned areas make it hard for the network to predict reliably the 3D scene. This in return degrades the localization accuracy. These results, when compared to DSAC++ and DSAC* (Tab. \ref{outdoor}) which choose few thousands of keypoints, infer that it is not the quantity of the correspondences but the quality of the correspondences what matters for accurate localization.

To complement this experiment, we consider a third set of 4800 correspondences (number of correspondences chosen by DSAC++ and DSAC*) and compute the runtime of the pose calculation step; i.e, the PnP in RANSAC framework. We use the OpenCV python implementation \cite{opencv_library}. Comparing set 1 to set 3, the runtime was improved by 14 times for OldHospital set and 15 times for KingsCollege set. With a python implementation using Pytorch \cite{pytorch}, one network forward pass to obtain the 2D Keypoints and the 3D coordinates needs around 8ms on single Nvidia GeForce GTX TITAN X. 

\begin{figure}[h]
\centering
\includegraphics[scale=0.3]{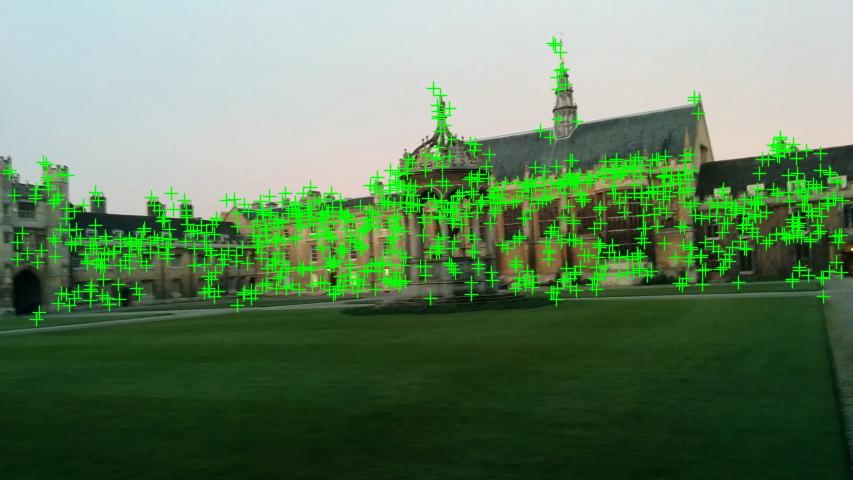}
  \caption{Keypoints selected from the GreatCourt landmark. The grass, the buildings roofs, and the sky are avoided for being non discriminatory thus not relevant for localization.}
  \label{exception}
\end{figure}

%\vspace{-0.7cm}
\subsection{Observations} 
We observed by training on Great Court landmark of Cambridge Dataset \cite{posenet} that the network could not learn the 3D coordinates. To our understanding, this could be because the 3D scene is far from the camera. Though this limitation, the network learned to avoid 2D keypoints selection from the grass spaces and the sky as can be seen in Fig. \ref{exception}. Similarly, training on Aachen dataset did not converge. A solution for this could be to cluster the scene and learn each division separately. On a different perspective and for general cases, where the number of selected correspondences is very low, the threshold can be lowered to select more keypoints and to avoid localization failure cases.
%\vspace{-0.5cm}
\section{CONCLUSIONS}
We have presented a method for camera 6DoF global pose estimation from a single RGB image that exploits discriminatory image regions to mitigate outliers and localize accurately and efficiently. Our method learns to carefully choose good features and avoid occlusions and other unreliable regions such as sky, streets, trees, bushes, pedestrians, and cars. We show that by avoiding these regions, outliers are minimized which allows the selection of a low number of correspondences to estimate a pose improving not only the localization accuracy but also the efficiency. To the best of our knowledge, our work is the first to exploit the concept of triangulated features from Structure from motion method to directly learn reliable and discriminatory image regions and the first to learn the 3D scene estimation and 2D keypoints detection in one framework. All of this combined has led our work to surpass in terms of localization accuracy and efficiency other learning-based and feature matching/aligning-based methods on scenes from Cambridge Landmarks dataset. We aim to explore directions for scaling up the localization scope of this work by for example pairing the work with an image retrieval system. We aim further to explore the potential of our keypoints detector for matching based localization approaches by forexample pairing it with learned or hand-crafted descriptors.

%%%%%%%%%%%%%%%%%%%%%%%%%%%%%%%%%%%%%%%%%%%%%%%%%%%%%%%%%%%%%%%%%%%%%%%%%%%%%%%%

%%%%%%%%%%%%%%%%%%%%%%%%%%%%%%%%%%%%%%%%%%%%%%%%%%%%%%%%%%%%%%%%%%%%%%%%%%%%%%%%

%%%%%%%%%%%%%%%%%%%%%%%%%%%%%%%%%%%%%%%%%%%%%%%%%%%%%%%%%%%%%%%%%%%%%%%%%%%%%%%%

%\section*{ACKNOWLEDGMENT}

%%%%%%%%%%%%%%%%%%%%%%%%%%%%%%%%%%%%%%%%%%%%%%%%%%%%%%%%%%%%%%%%%%%%%%%%%%%%%%%%

%\include{IEEexample.bib}

\bibliographystyle{IEEEtran}
\bibliography{IEEEabrv,root}

%\begin{thebibliography}{99}

%\bibitem{c3} H. Poor, An Introduction to Signal Detection and Estimation.   New York: Springer-Verlag, 1985, ch. 4.
%\end{thebibliography}

\end{document}